\documentclass[a4paper,11pt]{amsart}
\usepackage{amsmath,amsfonts,amssymb,amsthm, comment}
\usepackage{mathrsfs}
\usepackage{mathscinet}
\usepackage{graphicx}
\usepackage{enumerate}
\usepackage{algorithm}
\usepackage[noend]{algpseudocode}
\usepackage{color}
\usepackage[numbers]{natbib}
\usepackage[hyperindex,breaklinks]{hyperref}
\usepackage{xspace}
\usepackage{multirow}
\usepackage[noend]{algpseudocode}

\def\citeapos#1{\citeauthor{#1} (\citeyear{#1})}
\def\BState{\State\hskip-\ALG@thistlm}
\hypersetup{
  colorlinks   = true,   urlcolor     = blue,   linkcolor    = blue,   citecolor    = blue }
\providecommand{\U}[1]{\protect\rule{.1in}{.1in}}
\hypersetup{
pdftitle={},
pdfsubject={Probability Theory},
pdfauthor={Jose Blanchet, Karthyek Murthy},
pdfdisplaydoctitle=true,
}
\providecommand{\U}[1]{\protect\rule{.1in}{.1in}}

\theoremstyle{remark}

\begin{document}
\title[]{Doubly Robust Data-Driven Distributionally Robust Optimization}
\author{Blanchet, J.}
\address{Columbia University, Department of Statistics and Department of
Industrial Engineering \& Operations Research, New York, NY 10027, United
States.}
\email{jose.blanchet@columbia.edu}
\author{Kang, Y.}
\address{Columbia University, Department of Statistics. New York, NY 10027,
United States.}
\email{yangkang@stat.columbia.edu}
\author{Zhang, F.}
\address{Columbia University, Department of Statistics and Department of
Industrial Engineering \& Operations Research. New York, NY 10027, United
States.}
\email{fz2222@columbia.edu }
\author{He, F.}
\address{Columbia University, Department of Statistics and Department of
Industrial Engineering \& Operations Research. New York, NY 10027, United
States.}
\email{fh2293@columbia.edu }
\author{Hu, Z.}
\email{hu.zhangyi@gmail.com }
\date{\today }

\begin{abstract}
  Data-driven Distributionally Robust Optimization (DD-DRO) via optimal transport
  has been shown to encompass a wide range of popular machine learning algorithms. The distributional uncertainty
  size is often shown to correspond to the regularization parameter. 
  The type of regularization (e.g. the norm used to regularize) corresponds to 
  the shape of the distributional uncertainty. We propose a data-driven robust optimization
  methodology to inform the transportation cost underlying the definition of the distributional uncertainty.
  We show empirically that this additional layer of robustification, which produces a method we called
  doubly robust data-driven distributionally robust optimization (DD-R-DRO), allows to enhance the generalization properties of regularized estimators while
  reducing testing error relative to state-of-the-art classifiers in a wide range of data sets. 
\end{abstract}

\maketitle
\section{Introduction}

A wide class of popular machine learning estimators have been recently shown
to be particular cases of data-driven Distributionally Robust Optimization
(DD-DRO) formulations with a distributional uncertainty set centered around
the empirical distribution.
\smallskip\newline
For example, regularized logistic regression, support vector machines and
sqrt-Lasso, among many other machine learning formulations can be exactly
represented as DD-DRO problems involving an uncertainty set comprised of
probability distributions which are within a distance $\delta $ from the
empirical distribution. The distance is measured in terms of a class of
suitably defined Wasserstein distances or, more generally, optimal transport
distances between distributions.
\smallskip\newline
Our contribution in this paper is to build an additional robustification
layer on top of the DD-DRO formulation which encompasses the machine
learning algorithms mentioned earlier. Because of the second layer of
robustification, we call our approach DD-R-DRO. 
\smallskip\newline
More specifically, we consider a parametric family of optimal transport distances
and formulate a data-driven Robust Optimization (RO) problem for the
selection of such a\ distance, which in turn is used to inform the
distributional uncertainty region in the type of DD-DRO mentioned in the
previous paragraph. In addition, we provide an iterative algorithm for
solving such RO problem.
\smallskip\newline
In order to explain DD-R-DRO more precisely, let us discuss different
layers of robustness that are added in our various optimization formulations
and how these layers translate in terms of machine learning properties.
\smallskip\newline
A DD-DRO problem takes the general form 
\begin{equation}
\min_{\beta }\max_{P\in \mathcal{U}_{\delta }\left( P_{n}\right) }\mathbb{E}_{P}\left[
l\left( X, Y,\beta \right) \right] ,  \label{Eqn-DRO_origin}
\end{equation}%
where $\beta $ is a decision variable, $(X,Y)$ is a random element, and $l\left(
X,Y,\beta \right) $ is a loss incurred if the decision $\beta $ is taken and $%
(X, Y) $ is realized. The expectation $\mathbb{E}_{P}[\cdot]$ is taken under
the probability model $P$. The set $\mathcal{U}_{\delta }\left( P_{n}\right) 
$ is called the distributional uncertainty set; it is centered around the
empirical distribution $P_{n}$ of the data, and it is indexed by the
parameter $\delta >0$, which measures the size of the distributional
uncertainty.
\smallskip\newline
The min-max problem in (\ref{Eqn-DRO_origin}) can be interpreted as a game.
We (the outer player) wish to learn a task using a class of machines indexed
by $\beta $. An adversary (the inner player) is introduced to enhance
out-of-sample performance. The adversary has a budget $\delta $ and can
perturb the data, represented by $P_{n}$, in a certain way -- this is
important and we will return to this point. By introducing the artificial
adversary and the distributional uncertainty, the DD-DRO formulation provides
a direct mechanism to control the generalization properties of the learning
procedure.
\smallskip\newline
To further connect the DD-DRO representation (\ref{Eqn-DRO_origin}) with
more mainstream machine learning mechanisms for the control of out-of-sample
performance (such as regularization), we recall one of the explicit
representations given in \citeapos{blanchet2016robust}.
\smallskip\newline
In the context of generalized logistic regression (i.e. if the $l\left(
x,y,\beta \right) =\log \left( 1+\exp \left( -y\beta ^{T}x\right) \right) $%
), given an empirical sample $\mathcal{D}_{n}=\left\{ \left(
X_{i},Y_{i}\right) \right\} _{i=1}^{n}$ with $Y_{i}\in \{-1,1\}$ and a
judicious choice of the distributional uncertainty $\mathcal{U}_{\delta
}\left( P_{n}\right) $, \citeapos{blanchet2016robust} shows that%
\begin{equation}
\min_{\beta }\max_{P\in \mathcal{U}_{\delta }\left( P_{n}\right) }\mathbb{E}%
_{P}[l\left( X,Y,\beta \right) ]=\min_{\beta }\left( \mathbb{E}%
_{P_{n}}[l\left( X,Y,\beta \right) ]+\delta \left\Vert \beta \right\Vert
_{p}\right) ,  \label{DR_Las}
\end{equation}%
where $\left\Vert \cdot \right\Vert _{p}$ is the $l_{p}$ norm in $\mathbb{R}%
^{d}$ for $p\in \lbrack 1,\infty )$ and $\mathbb{E}_{P_{n}}[l\left( X,Y,\beta \right) ]=n^{-1}\sum_{i=1}^{n}l\left(
X_{i},Y_{i},\beta \right)$.
\smallskip\newline
The definition of $\mathcal{U}_{\delta }\left( P_{n}\right) $ turns out to
be informed by the dual norm $\left\Vert \cdot \right\Vert _{q}$ with $%
1/p+1/q=1$. In simple words, the \textit{shape} of the distributional
uncertainty $\mathcal{U}_{\delta }\left( P_{n}\right) $ directly implies the 
\textit{type of regularization}; and the \textit{size} of the distributional
uncertainty, $\delta $, dictates the regularization parameter.
\smallskip\newline
The story behind the connection to sqrt-Lasso, support vector machines and
other estimators is completely analogous to that given for (\ref{DR_Las}). A
key point in most of the known representations, such as (\ref{DR_Las}), is
that they are only partially informed by data. Only the center, $P_{n}$, and
the size, $\delta $ (via cross validation) are informed by data, but not the
shape.
\smallskip\newline
In recent work, \citeapos{blanchet2017data} proposes using metric learning procedures to
inform the shape of the distributional uncertainty. But the procedure
proposed in \citeapos{blanchet2017data}, though data-driven, is not robustified.
\smallskip\newline
One of the driving points of using robust optimization techniques in machine
learning is that the introduction of an adversary can be seen as a tool to
control the testing error. While the data-driven procedure in \citeapos{blanchet2017data}
is rich in the use of information, and hence it is able to improve the
generalization performance, the lack of robustifiation exposes the testing
error to potentially high variability. So, our contribution in this paper is
to design an RO procedure for choosing the shape of $\mathcal{U}_{\delta
}\left( P_{n}\right) $ using a suitable parametric family. In the context of
logistic regression, for example, the parametric family that we consider
includes the type of choice leading to (\ref{DR_Las}) as a particular case.
In turn, the choice of $\mathcal{U}_{\delta }\left( P_{n}\right) $ is
applied to formulation (\ref{Eqn-DRO_origin}) in order to obtain a \textit{%
	doubly-robustified} estimator.
 \begin{figure}[th]
 	\vskip 0.2in
 	\par
 	\begin{center}
 		\centerline{\includegraphics[width=16.5cm]{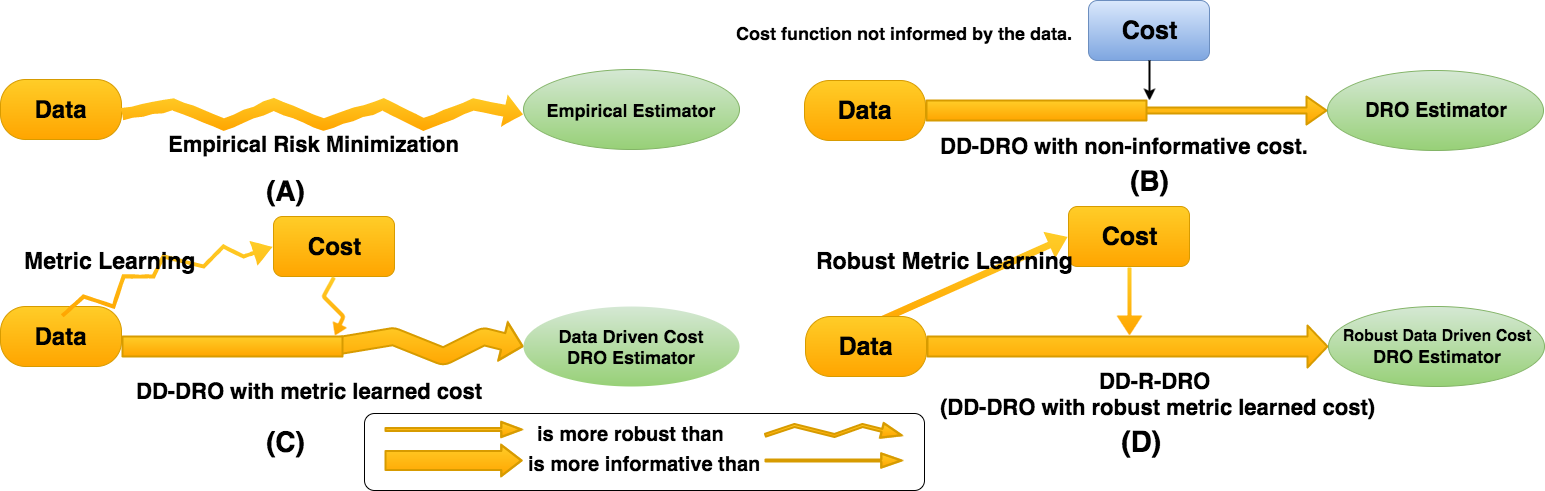}}
 	\end{center}
 	\par
 	\vskip -0.2in\caption{Four diagrams illustrating information on robustness.}%
 	\label{Fig_DRO_intui0}%
 \end{figure}
\smallskip\newline

Figure \ref{Fig_DRO_intui0} shows the various combinations of information and robustness which
have been studied in the literature so far. The figure shows four diagrams.
Diagram (A) represents standard empirical risk minimization (ERM);
which fully uses the information but often leads to high variability in testing
error and, therefore, poor out-of-sample performance. Diagram (B)
represents DD-DRO where only the center, $P_{n}$, and the size of the
uncertainty, $\delta $, are data driven; this choice controls out-of-sample
performance but does not use data to shape the type of perturbation, thus
potentially resulting in testing error bounds which might be pessimistic.
Diagram (C) represents DD-DRO\ with data-driven shape information for
perturbation type using metric learning techniques; this construction can
reduce the testing error bounds at the expense of increase in the
variability of the testing error estimates. Diagram (D) represents
DD-R-DRO, the shape of the perturbation allowed for the adversary
player is estimated using an RO procedure; this double robustification, as
we shall show in the numerical experiments is able to control the
variability present in the third diagram.  
\smallskip\newline
In the diagrams, the straight arrows represent the use of a robustification
procedure. A wide arrow represents the use of high degree of information. A
wiggly arrow indicates potentially noisy testing error estimates. 
\smallskip\newline
The contributions of this paper can be stated, in order of importance, as
follows:
\smallskip\newline
1) The fourth diagram, DD-R-DRO, illustrates the main contribution of this
paper, namely, a double robustification approach which reduces the
generalization error, utilizes information efficiently, and controls
variability.
\newline
2) An explicit RO formulation for metric learning tasks.
\newline
3)\ Iterative procedures for the solution of these RO problems.

\section{DD-DRO, Optimal Transport, and Machine Learning}

Let us consider a supervised machine learning classification
problem, where we have a response $Y\in \{-1,1\}$ and predictors $X\in 
\mathbb{R}^{d}$. Underlying there is a general loss function $l(x,y,\beta)$ and a class of classifiers indexed by the parameter $\beta$. The distributional uncertainty region in (\ref{Eqn-DRO_origin}) takes the form
\begin{equation}
\mathcal{U}_{\delta }\left( P_{n}\right) =\left\{ P:D_{c}\left(
P,P_{n}\right) \leq \delta \right\} ,  \label{U_delta}
\end{equation}%
where $D_{c}\left( P,P_{n}\right) $ is a suitably defined notion of
discrepancy between $P$ and $P_{n}$ so that $D_{c}\left( P,P_{n}\right) =0$
implies that $P=P_{n}$.
\smallskip\newline
Other notions of discrepancy have been considered in the DRO literaturem for example the Kullback-Leibler
divergence (or another divergence notion which depends on the likelihood
ratio) is utilized \citeapos{hu2013kullback}. Unfortunately, divergence criteria which
relies on the existence of the likelihood ratio between $P$ and $P_{n}$
ultimately forces $P$ to share the same support as $P_{n}$, therefore
potentially inducing undesirable out-of-sample performance.
\smallskip\newline
Instead, we follow the approach in \citeapos{esfahani2015data}, \citeapos{shafieezadeh2015distributionally}, and \citeapos{blanchet2016robust}, and
define $D_{c}\left( P,P_{n}\right) $ as the optimal transport discrepancy
between $P$ and $P_{n}$.

\subsection{Optimal Transport Distances and Discrepancies}

Assume that the cost function $c:\mathbb{R}^{d+1}\times \mathbb{R}%
^{d+1}\rightarrow \lbrack 0,\infty ]$ is lower semicontinuous. We also
assume that $c(u,v)=0$ if and only if $u=v$.
\smallskip\newline
Given two distributions $P$ and $Q$, with supports $\mathcal{S}_{P}$ and $%
\mathcal{S}_{Q}$, respectively, we define the optimal transport discrepancy, 
$D_{c}$, via%
\begin{equation}\label{Discrepancy_Def}
D_{c}\left( P,Q\right) =\inf \{E_{\pi }\left[ c\left( U,V\right) \right]
:\pi \in \mathcal{P}\left( \mathcal{S}_{P}\times \mathcal{S}_{Q}\right) ,%
\text{ }\pi _{U}=P,\text{ }\pi _{V}=Q\},
\end{equation}%
where $\mathcal{P}\left( \mathcal{S}_{P}\times \mathcal{S}_{Q}\right) $ is
the set of probability distributions $\pi $ supported on $\mathcal{S}%
_{P}\times \mathcal{S}_{Q}$, and $\pi _{U}$ and $\pi _{V}$ denote the
marginals of $U$ and $V$ under $\pi $, respectively. Because $c\left( \cdot
\right) $ is non-negative we have that $D_{c}\left( P,Q\right) \geq 0$.
Moreover, requiring that $c\left( u,v\right) =0$ if and only if $u=v$
guarantees that $D_{c}\left( P,Q\right) =0$ if and only $P=Q$. 
\smallskip\newline
If, in addition, $c\left( \cdot \right) $ is symmetric (i.e. $c\left(
u,v\right) =c\left( v,u\right) $), and there exists $\varrho \geq 1$ such
that $c^{1/\varrho }\left( u,w\right) \leq c^{1/\varrho }\left( u,v\right)
+c^{1/\varrho }\left( v,w\right) $ (i.e. $c^{1/\varrho }\left( \cdot \right) 
$ satisfies the triangle inequality) then it can be easily verified (see %
\citeapos{villani2008optimal}) that $%
D_{c}^{1/\varrho }\left( P,Q\right) $ is a metric. For example, if $c\left(
u,v\right) =\left\Vert u-v\right\Vert _{q}^{\varrho }$ for $q\geq 1$ (where $%
\left\Vert u-v\right\Vert _{q}$ denotes the $l_{q}$ norm in $\mathbb{R}%
^{d+1} $) then $D_{c}\left( \cdot \right) $ is known at the Wasserstein
distance of order $\varrho $. 
\smallskip\newline
Observe that (\ref{Discrepancy_Def}) is obtained by solving a linear
programming problem. For example, suppose that $Q=P_{n}$, so $Q\in \mathcal{P%
}\left( \mathcal{D}_{n}\right) $ and assume that the support $\mathcal{S}%
_{P} $ of $P$ is finite. Then, using $U=\left( X,Y\right) $, we have that $%
D_{c}\left( P,P_{n}\right) $ is obtained by computing%
\begin{align}  \label{LP}
	&\min_{\pi}\sum_{u\in \mathcal{S}_{P}} \sum_{v\in \mathcal{D}_{n}}c\left( u,v\right)\pi \left( u,v\right): \\\text{s.t.}&
	\sum_{u\in \mathcal{S}_{P}}\pi \left( u,v\right) =\frac{1}{n}\text{ 
	}\forall \text{ }v\in \mathcal{D}_{n}\nonumber\\
	&\sum_{v\in \mathcal{D}_{N}}\pi \left( u,v\right) =P\left( \left\{
	u\right\} \right) \text{ }\forall \text{ }u\in \mathcal{X}_{N},\nonumber\\
	& \pi \left( u,v\right) \geq 0\text{ }\forall \text{ }\left(
	u,v\right) \in \mathcal{S}_{P}\times \mathcal{D}_{n}\nonumber
\end{align}

A completely analogous linear program (LP), albeit an infinite dimensional
one, can be defined if $\mathcal{S}_{P}$ has infinitely many elements. This
LP has been extensively studied in great generality in the context of
Optimal Transport under the name of Kantorovich's problem (see \citeapos%
{villani2008optimal})). Requiring $c\left( \cdot \right) $ to be lower
semicontinuous guarantees the existence of an optimal solution to
Kantorovich's problem.
\smallskip\newline
Note that $D_{c}\left( P,P_{n}\right) $ can be interpreted as the minimal
cost of rearranging (i.e. transporting the mass of) the distribution $P_{n}$
into the distribution $P$. The rearrangement mechanism has a transportation
cost $c\left( u,w\right) \geq 0$ for moving a unit of mass from location $u$
in the support of $P_{n}$ to location $w$ in the support of $P$. For
instance, in the setting of (\ref{DR_Las}) we have that 
\begin{equation}
c\left( \left( x,y\right) ,\left( x^{\prime },y^{\prime }\right) \right)
=\left\Vert x-x^{\prime }\right\Vert _{q}^{2}I\left( y=y^{\prime }\right)
+\infty \cdot I\left( y\neq y^{\prime }\right) .  \label{Cost}
\end{equation}

The infinite contribution in the definition of $c$ (i.e. $\infty \cdot
I\left( y\neq y^{\prime }\right) $) indicates that the adversary player in
the DRO formulation is not allowed to perturb the response variable.

\section{Data-Driven Selection of Optimal Transport Cost Function}

By suitably choosing $c\left( \cdot \right) $ we might further improve the
generalization properties of the DD-DRO estimator based on (\ref%
{Eqn-DRO_origin}). To fix ideas, consider a suitably parameterized family of
transportation costs as follows. Let $\Lambda $ be a positive semidefinite
matrix (denoted as $\Lambda \succeq 0$) and define $\left\Vert x\right\Vert
_{\Lambda }^{2}=x^{T}\Lambda x$. Inspired by (\ref{Cost}), consider the cost
function%
\begin{equation}
c_{\Lambda }\left( \left( x,y\right) ,\left( x^{\prime },y^{\prime }\right)
\right) =d_{\Lambda }^{2}\left( x,x^{\prime }\right) I\left( y=y^{\prime
}\right) +\infty I\left( y\neq y^{\prime }\right) ,  \label{c_lambda}
\end{equation}%
where $d_{\Lambda }^{2}\left( x,x^{\prime }\right) =\left\Vert x-x^{\prime
}\right\Vert _{\Lambda }^{2}$. Then, \citeapos{blanchet2017data} shows that in the
generalized logistic regression setting (i.e. $l\left( x,y,\beta \right)
=\log (1+\exp \left( -y\beta ^{T}x\right) )$), if $\Lambda $ is positive
definite, we obtain%
\begin{equation}
\min_{\beta }\max_{P:D_{c_{\Lambda }}\left( P,P_{n}\right) \leq \delta }%
\mathbb{E}\left[ l\left( X,Y,\beta \right) \right] =\min_{\beta }\mathbb{E}%
_{P_{n}}\left[ l\left( X,Y,\beta \right) \right] +\delta \left\Vert \beta
\right\Vert _{\Lambda ^{-1}}.  \label{GAR_2}
\end{equation}

If the choice of $\Lambda $ is data driven in order to impose a penalty on
transportation costs whose outcomes that are highly impactful in terms of
risk, then we would be able to control the risk bound induced by the DD-DRO
formulation. This is the strategy studied in \citeapos{blanchet2017data} in which metric
learning procedures have been implemented precisely to achieve such control.
Our contribution, as we shall explain in the next section is the use of a
robust optimization formulation to calibrate $c_{\Lambda }\left( \cdot
\right) $. We emphasize that once $c_{\Lambda }\left( \cdot \right) $ is
calibrated, it can be used to multiple learning tasks and arbitrary loss
functions (not only logistic regression).

\subsection{Data-Driven Cost Functions via Metric Learning Procedures\label%
	{Sec-Robust-DD-Cost}}

We quickly review the elements of standard metric learning procedures. Our
data is of the form $\mathcal{D}_{n}=\left\{ (X_{i},Y_{i})\right\}
_{i=1}^{n} $ and $Y_{i}\in \{-1,+1\}$. The prediction variables are assumed
to be standardized.
\smallskip\newline
Motivated by applications such as social networks, in which there is a
natural graph which can be used to connect instances in the data, we assume
that one is given sets $\mathcal{M}$ and $\mathcal{N}$, where $\mathcal{M}$
is the set of the pairs that should be close (so that we can connect them)
to each other, and $\mathcal{N}$, on contrary, is characterizing the
relations that the pairs should be far away (not connected), we define them
as 
\begin{align*}
	\mathcal{M} &=\left\{ \left( X_{i},X_{j}\right) |X_{i}\text{
		and }X_{j}\text{ must connect}\right\} ,\\ \text{ and }
	\mathcal{N} &=\left\{ \left( X_{i},X_{j}\right) | X_{i}\text{
		and }X_{j}\text{ should not connect}\right\} .
\end{align*}

While it is typically assumed that $\mathcal{M}$ and $\mathcal{N}$ are
given, one may always resort to $k$-Nearest-Neighbor ($k$-NN) method for the
generation of these sets. This is the approach that we follow in our
numerical experiments. But we emphasize that choosing any criterion for the
definition of $\mathcal{M}$ and $\mathcal{N}$ should be influenced by the
learning task in order to retain both interpretability and performance. In
our experiments we let $\left( X_{i},X_{j}\right) $ belong to $\mathcal{M}$
if, in addition to being sufficiently close (i.e. in the $k$-NN criterion), $%
Y_{i}=Y_{j}$. If $Y_{i}\neq Y_{j}$, then we have that $\left(
X_{i},X_{j}\right) \in \mathcal{N}$.
\smallskip\newline
In addition, we consider the relative constraint set $\mathcal{R}$
containing data triplets with relative relation defined as 
\begin{equation*}
\mathcal{R}=\left\{ \left( i,j,k\right) |d_{\Lambda }(X_{i},X_{j})\text{
	should be smaller than }d_{\Lambda }(X_{i},X_{k})\right\} .
\end{equation*}

Let us consider the following two formulations of metric learning, the
so-called Absolute Metric Learning formulation 
\begin{equation}
\min_{\Lambda \succeq 0}\sum_{\left( i,j\right) \in \mathcal{M}}d_{\Lambda
}^{2}(X_{i},X_{j})\quad \text{s.t.}\quad \sum_{\left( i,j\right) \in 
\mathcal{N}}d_{A}^{2}(X_{i},X_{j})\geq 1,  \label{Eqn-Metric-Learn-Opt}
\end{equation}%
and the Relative Metric Learning formulation, 
\begin{equation}
\min_{\Lambda \succeq 0}\sum_{\left( i,j,k\right) \in \mathcal{R}}\left(
d_{\Lambda }^{2}\left( X_{i},X_{j}\right) -d_{\Lambda }^{2}\left(
X_{i},X_{k}\right) +1\right) _{+}.  \label{Eqn-ML-Relative}
\end{equation}

Both formulations have their merits, \eqref{Eqn-Metric-Learn-Opt} exploits
both the constraint sets $\mathcal{M}$ and $\mathcal{N}$, while %
\eqref{Eqn-ML-Relative} is only based on information in $\mathcal{R}$.
Further intuition or motivation of those two formulations can be found in 
\citeapos{xing2002distance} and \citeapos{weinberger2009distance}, respectively. We
will show how to formulate and solve the robust counterpart of those two
representative examples by robustifying a single constraint set or two sets
simultaneously For simplicity we only discuss these two formulations, but
many metric learning algorithms are based on natural generalizations of
those two forms, as mentioned in the survey \citeapos{bellet2013survey}.
\smallskip\newline
Once formulation (\ref{Eqn-Metric-Learn-Opt}) or (\ref{Eqn-ML-Relative}) are
considered and the matrix $\Lambda $ has been calibrated, one may then
consider the cost function $c_{\Lambda }\left( \cdot \right) $ in (\ref%
{c_lambda}) and solve the problem (\ref{GAR_2}). This is the benchmark that
we will consider in our numerical experiments. And we will contrast this
approach versus a method which chooses $\Lambda $ using a robust
optimization version of (\ref{Eqn-Metric-Learn-Opt}) or (\ref%
{Eqn-ML-Relative}) as we shall explain next.

\section{Robust Optimization for Metric Learning}

In this section, we review a robust optimization method to metric learning
optimization problem to learn a robust data-driven cost function. RO is a
family of optimization techniques that deals with uncertainty or
misspecification in the objective function and constraints. RO was first
proposed in \citeapos{ben2009robust} and has attracted increasing attentions in
the recent decades \citeapos{el1997robust} and \citeapos{bertsimas2011theory}. RO has
been applied in machine learning to regularize statistical learning
procedures, for example, in \citeapos{xu2009robust} and \citeapos{xu2009robustness}
robust optimization was employed for SR-Lasso and support vector machines.
We apply RO, as we shall demonstrate, to reduce the variability in testing
error when implementing DD-DRO.

\subsection{Robust Optimization for Relative Metric Learning}

The RO formulation that we shall use for \eqref{Eqn-ML-Relative} is based on
the work of \citeapos{huang2012robust}. In order to motivate this formulation, suppose that we
know that only $\alpha $ level, e.g. $\alpha =90\%$, of the constraints are
satisfied, but we do not have information on exactly which of them are
ultimately satisfied. The value of $\alpha $ may be inferred using cross
validation.
\smallskip\newline
Instead of optimizing over all subsets of constraints, we try to minimize
the worst case loss function over all possible $\alpha \left\vert \mathcal{R}%
\right\vert $ constraints (where $\left\vert \cdot \right\vert $ is
cardinality of a set) and obtain the following min-max formulation 
\begin{equation}
\min_{\Lambda \succeq 0}\max_{\tilde{q}\in \mathcal{T}(\alpha
	)}\sum_{(i,j,k)\in \mathcal{R}}q_{i,j,k}\left( d_{\Lambda }^{2}\left(
X_{i},X_{j}\right) -d_{\Lambda }^{2}\left( X_{i},X_{k}\right) + 1\right) _{+},
\label{Eqn-ML-relative-robust}
\end{equation}%
where $\mathcal{T}(\alpha )$ is a robust uncertainty set of the form 
\[
\mathcal{T}\left( \alpha \right) =\big\{ \tilde{q}=\left\{
q_{i,j,k}|(i,j,k)\in \mathcal{R}\right\} |0\leq q_{i,j,k}\leq
1,\sum_{(i,j,k)\in \mathcal{R}}q_{i,j,k}\leq \alpha \times \left\vert 
\mathcal{R}\right\vert \big\} ,
\]
which is a convex and compact set.
\smallskip\newline
In addition, the objective function in \eqref{Eqn-ML-Relative} is convex in $%
\Lambda $ and concave (linear) in $\tilde{q}$, so we can switch the order of
min-max by resorting to Sion's min-max theorem (\citeapos{terkelsen1973some}).
This important observation suggests an iterative algorithm. For a fixed $%
\Lambda \succeq 0$, the inner maximization is linear in $\tilde{q}$, and the
optimal $\tilde{q}$ satisfy $\tilde{q}_{i,j,k}=1$ whenever $\left(
d_{\Lambda }\left( X_{i},X_{j}\right) -d_{\Lambda }\left( X_{i},X_{k}\right)
+1\right) _{+}$ ranks in the top $\alpha \left\vert \mathcal{R}\right\vert $
largest values and equals $\tilde{q}_{i,j,k}$ otherwise.
\smallskip\newline
Let us use $\mathcal{R}_{\alpha }\left( \Lambda \right) $ to denote the
subset of constraints satisfying that the corresponding loss function $%
\left( d_{\Lambda }\left( X_{i},X_{j}\right) -d_{\Lambda }\left(
X_{i},X_{k}\right) +1\right) _{+}$ ranks at the top $\alpha |\mathcal{R}|$
largest values among the corresponding loss function values of the triplets
in $\mathcal{R}$.
\smallskip\newline
For fixed $\tilde{q}$, the optimization problem is convex in $\Lambda $, we
can solve this problem using sub-gradient or smoothing approximation
algorithms (\citeapos{nesterov2005smooth}). Particularly, as we discussed above,
if $\tilde{q}$ is the solution for fixed $\Lambda $, we know, solving $%
\Lambda $ is equivalent to solving its non-robust counterpart %
\eqref{Eqn-ML-Relative}, replacing $\mathcal{R}$ by $\mathcal{R}_{\alpha
}(\Lambda )$, where $\mathcal{R}_{\alpha }(\Lambda )$ is a subset of $%
\mathcal{R}$ that contains the constraints have top $\alpha |\mathcal{R}|$
violation, i.e. 
\begin{equation*}
\mathcal{R}_{\alpha }\left( \Lambda \right) =\left\{ (i,j,k)\in \mathcal{R}%
|\left( d_{\Lambda }^{2}\left( X_{i},X_{j}\right) -d_{\Lambda }^{2}\left(
X_{i},X_{k}\right) +1\right) _{+}\text{ ranks top }\alpha \text{ within }%
\mathcal{R}\right\} .
\end{equation*}%
We summarize the sub-gradient based sequentially update algorithm as in
Algorithm \ref{Algo-Metric-Robust-Seq-Relative}.

\begin{algorithm}
	\caption{Sequential Coordinate-wise Metric Learning Using Relative Relations} \label{Algo-Metric-Robust-Seq-Relative}
	\begin{algorithmic}[1]
		\State \textbf{Initialize} $\Lambda = I_{d}$, learning rate $\alpha = 0.01$ tracking error $Error=1000$ as a large number.	Then randomly sample $\alpha$ proportion of elements from $\mathcal{R}$ to construct $\mathcal{R}_{\alpha}(\Lambda)$.		
		\While {$Error > 10^{-3}$}
		\State Update $\Lambda$ using projected (projected to positive semidefinite matrix cone) subgradient descent technique.
		\begin{align*}
		\Lambda = \pi_{\mathbb{S_{+}}}
		\big(\Lambda - \alpha\sum_{\left(i,j,k\right)\in\mathcal{R}_{\alpha}\left(\Lambda\right)}
		\nabla_{\Lambda}\left(d_{\Lambda}^{2}\left(X_i,X_j\right) 
		- d_{\Lambda}^{2}\left(X_i,X_k\right)+1\right)_{+}\big)
		\end{align*}
		\State Update tracking error $Error$ as the norm of difference between latest matrix $\Lambda$ and average of last $50$ iterations.
		\State Every few steps (5 or 10 iterations), compute $\left(d_{\Lambda}^{2}\left(X_i,X_j\right) - d_{\Lambda}^{2}\left(X_i,X_k\right)+1\right)_{+}$ for all $\left(i,j,k\right)\in \mathcal{R}$, then update $\mathcal{R}_{\alpha}(\Lambda)$.
		\EndWhile
		\State \textbf{Output} $\Lambda$.		
	\end{algorithmic}
\end{algorithm}

As a remark, we would like to highlight the following
observations. While we focus on metric learning simply as a loss
minimization procedure as in \eqref{Eqn-ML-Relative} and
\eqref{Eqn-ML-relative-robust} for simplicity, in practice 
people usually add a
regularization term (such as $\left\Vert \Lambda\right\Vert_{F}$) to
the loss minimization, as is common in metric learning
literature (see \citeapos{bellet2013survey}). It is easy to observe
 our discussion above regarding the min-max exchange uses Sion's min-max
  theorem and everything else remains largely intact if we consider regularization.
 Likewise, one can use a more general loss functions than
the hinge loss used in \eqref{Eqn-ML-Relative} and
\eqref{Eqn-ML-relative-robust}.


\subsection{Robust Optimization for Absolute Metric Learning}

The RO formulation that we present here for \eqref{Eqn-Metric-Learn-Opt}
appears to be novel in the literature. Note that \eqref{Eqn-Metric-Learn-Opt}
can be written into the Lagrangian form, 
\begin{equation*}
\min_{\Lambda \succeq 0}\quad \max_{\lambda \geq 0}\quad \sum_{\left(
	i,j\right) \in \mathcal{M}}d_{\Lambda }^{2}\left( X_{i},X_{j}\right)
+\lambda \big( 1-\sum_{\left( i,j\right) \in \mathcal{N}}d_{\Lambda
}^{2}\left( X_{i},X_{j}\right) \big) .
\end{equation*}

Following similar discussion for $\mathcal{R}$, let us assume that the sets $%
\mathcal{M}$ and $\mathcal{N}$ are noisy or inaccurate at level $\alpha $
(i.e. $\alpha \cdot 100\%$ of their elements are incorrectly assigned). We
can construct robust uncertainty sets $\mathcal{W}(\alpha )$ and $\mathcal{V}%
(\alpha )$ from the constraints in $\mathcal{M}$ and $\mathcal{N}$ as
follows, 
\begin{align*}
\mathcal{W}(\alpha )& =\big\{ \tilde{\eta}=\left\{ \eta _{ij}:(i,j)\in 
\mathcal{M}\right\} | 0\leq \eta _{ij}\leq 1,\sum_{(i,j)\in 
	\mathcal{M}}\eta _{ij}\leq \alpha \times |\mathcal{M}| \big\} , \\
\mathcal{V}(\alpha )& =\big\{ \tilde{\xi}=\left\{ \xi _{ij}:(i,j)\in 
\mathcal{N}\right\} | 0\leq \xi _{ij}\leq 1,\sum_{(i,j)\in \mathcal{%
		N}}\xi _{ij}\geq \alpha \times |\mathcal{N}| \big\} .
\end{align*}
Then we can write the RO counterpart for the loss minimization problem of
metric learning as 
\begin{equation}
\min_{\Lambda \succeq 0}\quad \max_{\lambda \geq 0}\quad \max_{\tilde{\eta}%
	\in \mathcal{W}(\alpha ),\tilde{\xi}\in \mathcal{V}(\alpha )}\sum_{(i,j)\in 
	\mathcal{M}}\eta _{i,j}d_{\Lambda }^{2}\left( X_{i},X_{j}\right) +\lambda
\big( 1-\sum_{(i,j)\in \mathcal{N}}\xi _{i,j}d_{\Lambda }^{2}\left(
X_{i},X_{j}\right) \big)   \label{Eqn-ML-Absolute-Robust}
\end{equation}%
Note that the Cartesian product $\mathcal{W}\left( \alpha \right) \times 
\mathcal{V}\left( \alpha \right) $ is a compact set, and the objective function is convex in $\Lambda$ and concave (linear) in pair $(\tilde{\eta},\tilde{%
	\xi})$, so we can apply Sion's min-max Theorem again (see in \citeapos%
{terkelsen1973some}) to switch the order of min$_{\Lambda }$-max$_{(\tilde{%
		\eta},\tilde{\xi})}$ (after switching max$_{\lambda }$ and max$_{(\tilde{\eta%
	},\tilde{\xi})}$, which can be done in general). Then we can develop a
sequential iterative algorithm to solve this problem as we describe next.
\smallskip\newline
At the $k$-th step, given fixed $\Lambda _{k-1}\succeq 0$ and $\lambda
_{k-1}>0$ (it is easy to observe that optimal solution $\lambda $ is
positive, i.e. the constraint is active so we may safely assume $\lambda
_{k-1}>0$), the inner maximization problem, becomes, 
\begin{equation*}
\max_{\tilde{\eta}\in \mathcal{W}\left( \alpha \right) }\sum_{(i,j)\in 
	\mathcal{M}}\eta _{i,j}d_{\Lambda _{k-1}}^{2}\left( X_{i},X_{j}\right)
+\lambda \big( 1-\min_{\tilde{\xi}\in \mathcal{V}\left( \alpha \right)
}\sum_{(i,j)\in \mathcal{N}}\xi _{i,j}d_{\Lambda _{k-1}}^{2}\left(
X_{i},X_{j}\right) \big) .
\end{equation*}

As we discussed for relative constraints case, the optimal solution for $%
\tilde{\eta}$ and $\tilde{\xi}$ is, $\tilde{\eta}_{i,j}$ is 1, if $%
d_{\Lambda _{k-1}}^{2}\left( X_{i},X_{j}\right) $ ranks top $\alpha $ within 
$\mathcal{M}$ and equals 0 otherwise; while, on the contrary, $\tilde{\xi}%
_{i,j}=1$ if $d_{\Lambda _{k-1}}^{2}\left( X_{i},X_{j}\right) $ ranks bottom 
$\alpha $ within $\mathcal{N}$ and equals 0 otherwise.
\smallskip\newline

Similar as $\mathcal{R}_{\alpha }(\Lambda )$, we can define $%
\mathcal{M}_{\alpha }(\Lambda _{k-1})$ ( $\mathcal{N}_{\alpha }(\Lambda
_{k-1})$) as subset of $\mathcal{M}$ ($\mathcal{N}$), which contains the
constraints with largest $\alpha $ percent of $d_{\Lambda
	_{k-1}}\left( \cdot \right) $ within in $\mathcal{M}$; 
and  $\mathcal{N}_{\alpha }(\Lambda
_{k-1})$ as subset of $\mathcal{N}$, which contains the
constraints with smallest $\alpha $ percent of $d_{\Lambda
	_{k-1}}\left( \cdot \right) $ within in $\mathcal{N}$.
As we observe that the optimal $\tilde{\eta}_{i,j}=1$  if $(i,j)\in \mathcal{M}_{\alpha }(\Lambda _{k-1})$ 
and $\tilde{\xi}_{i,j}=1$ if $(i,j)\in \mathcal{N%
}_{\alpha }(\Lambda _{k-1})$, thus for fixed $\tilde{\eta}$ and $\tilde{\xi}
$, we can write the optimization problem over $\Lambda $ in the constrained
case as  
\begin{equation*}
\min_{\Lambda \succeq 0}\quad \sum_{\left( i,j\right) \in \mathcal{M}%
	_{\alpha }(\Lambda _{k-1})}d_{\Lambda }^{2}\left( X_{i},X_{j}\right) \quad 
\text{ s.t. }\quad \sum_{\left( i,j\right) \in \mathcal{N}_{\alpha }(\Lambda
	_{k-1})}d_{\Lambda }^{2}\left( X_{i},X_{j}\right) \geq 1.
\end{equation*}

This formulation falls within the setting of the problem stated in %
\eqref{Eqn-Metric-Learn-Opt} and thus it can be solved by using techniques
discussed in \citeapos{xing2002distance}. We summarize the details in Algorithm %
\ref{Algo-Metric-Robust-Seq}. 
\begin{algorithm}
	\caption{Sequential Coordinate-wise Metric Learning Using Absolute Constraints} \label{Algo-Metric-Robust-Seq}
	\begin{algorithmic}[1]
		\State \textbf{Initialize} $A = I_{d}$, tracking error $\text{Error}=1000$ as a large number.	Then randomly sample $\alpha$ proportion of elements from $\mathcal{M}$ (resp. $\mathcal{N}$) to construct $\mathcal{M}_{\alpha}(A)$ (resp. $\mathcal{N}_{\alpha}(A)$).		
		\While {$Error > 10^{-3}$}
		\State Update $A$ using procedure provided in \citeapos{xing2002distance}.
		\State Update tracking error $Error$ as the norm of difference between latest matrix $A$ and average of last $50$ iterations.
		\State Every few steps (5 or 10 iterations), compute $d_{A}\left(W_{i},W_{j}\right)$ for all $\left(i,j\right)\in \mathcal{M}\cup\mathcal{N}$, then update $\mathcal{M}_{\alpha}(A)$ and $\mathcal{N}_{\alpha}(A)$.
		\EndWhile
		\State \textbf{Output} $A$.		
	\end{algorithmic}
\end{algorithm}

Other robust methods have also been considered in the metric learning
literature, see \citeapos{zha2009robust} and \citeapos{lim2013robust} although the connections
to RO are not fully exposed.

\section{Numerical Experiments}
We proceed to numerical experiments to verify the performance of our DD-R-DRO
method empirically using six binary classification real data sets from UCI machine learning data base \citeapos{Lichman:2013}. 
\smallskip\newline
We consider logistic regression (LR), regularized logistic regression (LRL1), 
DD-DRO with cost function learned using absolute constraints (DD-DRO (absolute))
and its $\alpha=50\%,90\%$ level of doubly robust DRO (DD-R-DRO (absolute));
 DD-DRO with cost function learned using relative constraints (DD-DRO (relative)) and its 
  $\alpha=50\%,90\%$ level of doubly robust DRO (DD-R-DRO (relative)). For each data and each experiment, we
  randomly split the data into training and testing and fit models on training
  set and evaluate on testing set. 
\smallskip\newline
We report the mean and standard deviation of training error, testing error,
and testing accuracy via $200$ independent experiments for each data sets,
and summarize the detailed results and data set information (including split
setting) in Table \ref{Tab-real}. 
\smallskip\newline
For solving the DD-DRO and DD-R-DRO problem, we apply the smoothing approximation
algorithm introduced in \citeapos{blanchet2017data} to solve the DRO problem directly, where the 
size of uncertainty $\delta$ is chosen via $5-$fold cross-validation. 
\begin{table}[th]
  \centering
    \caption{Numerical results for real data sets.}
  { 
\begin{tabular}{cc|c|c|c|c|c|c}
&  & BC & BN & QSAR & Magic & MB & SB \\ \hline
\multicolumn{1}{c|}{\multirow{3}{*}{LR}} & Train & $0\pm0$ & $.008\pm.003$ & 
$.026\pm.008$ & $.213\pm.153$ & $0\pm 0$ & $0 \pm 0$ \\ 
\multicolumn{1}{c|}{} & Test & $8.75\pm 4.75$ & $2.80\pm1.44$ & $35.5\pm 12.8
$ & $17.8\pm 6.77$ & $18.2\pm 10.0$ & $14.5\pm 9.04$ \\ 
\multicolumn{1}{c|}{} & Accur & $.762\pm.061$ & $.926\pm.048$ & $.701\pm .040
$ & $.668\pm.042$ & $.678\pm.059$ & $.789 \pm .035$ \\ \hline
\multicolumn{1}{c|}{\multirow{3}{*}{LRL1}} & Train & $.185\pm.123$ & $%
.080\pm.030$ & $.614\pm.038$ & $.548\pm.087$ & $.401\pm .167$ & $.470 \pm
.040$ \\ 
\multicolumn{1}{c|}{} & Test & $.428\pm.338$ & $.340\pm.228$ & $.755\pm.019$
& $.610\pm.050$ & $.910\pm.131$ & $.588 \pm .140$ \\ 
\multicolumn{1}{c|}{} & Accur & $.929\pm.023$ & $.930\pm.042$ & $.646\pm .036
$ & $.665\pm.045$ & $.717\pm.041$ & $.811 \pm .034$ \\ \hline
\multicolumn{1}{c|}{\multirow{3}{*}{\begin{tabular}[c]{@{}c@{}}DD-DRO\\
(absolute)\end{tabular}}} & Train & $.022\pm.019$ & $.197\pm.112$ & $%
.402\pm.039$ & $.469\pm.064$ & $.294\pm.046$ & $.166 \pm .031$ \\ 
\multicolumn{1}{c|}{} & Test & $.126\pm.034$ & $.275\pm .093$ & $.557\pm .023
$ & $.571\pm .043$ & $.613\pm.053$ & $.333 \pm .023$ \\ 
\multicolumn{1}{c|}{} & Accur & $.954\pm.015$ & $.919\pm.050$ & $%
.733\pm.0.026$ & $.727\pm.039$ & $.714 \pm .032$ & $.887 \pm .011$ \\ \hline
\multicolumn{1}{c|}{\multirow{3}{*}{\begin{tabular}[c]{@{}c@{}}DD-R-DRO\\
(absolute)\\ $\alpha = 90\%$\end{tabular}}} & Train & $.029\pm .013$ & $.078
\pm .031$ & $.397 \pm .036$ & $.420\pm .063$ & $.249 \pm .055$ & $.194 \pm
.031$ \\ 
\multicolumn{1}{c|}{} & Test & $.126 \pm .023$ & $.259 \pm .086$ & $.554 \pm
.019$ & $.561\pm .035$ & $.609 \pm .044$ & $.331 \pm .018$ \\ 
\multicolumn{1}{c|}{} & Accur & $.954\pm .012$ & $.910 \pm .042$ & $.736 \pm
.025$ & $.729\pm .032$ & $.709 \pm .025$ & $.890\pm .008$ \\ \hline
\multicolumn{1}{c|}{\multirow{3}{*}{\begin{tabular}[c]{@{}c@{}}DD-R-DRO\\
(absolute)\\ $\alpha = 50\%$\end{tabular}}} & Train & $.040 \pm .055$ & $%
.137 \pm .030$ & $.448 \pm .032$ & $.504\pm .041$ & $.351 \pm .048$ & $.166
\pm .030$ \\ 
\multicolumn{1}{c|}{} & Test & $.132 \pm .015$ & $.288 \pm .059$ & $.579 \pm
.017$ & $.590\pm .029$ & $.623 \pm .029$ & $.337 \pm .013$ \\ 
\multicolumn{1}{c|}{} & Accur & $.952\pm .012$ & $.918 \pm .037$ & $.733 \pm
.025$ & $.710 \pm .033$ & $.715 \pm .021$ & $.888 \pm .008$ \\ \hline
\multicolumn{1}{c|}{\multirow{3}{*}{\begin{tabular}[c]{@{}c@{}}DD-DRO \\
(relative)\end{tabular}}} & Train & $.086 \pm .038$ & $.436 \pm .138$ & $%
.392 \pm .040$ & $.457 \pm .071$ & $.322 \pm .061$ & $.181 \pm ,036$ \\ 
\multicolumn{1}{c|}{} & Test & $.153 \pm .060$ & $.329 \pm .124$ & $.559 \pm
.025$ & $582 \pm .033$ & $.613 \pm .031$ & $.332 \pm .016$ \\ 
\multicolumn{1}{c|}{} & Accur & $.946 \pm .018$ & $.916 \pm .075$ & $.714
\pm .029$ & $.710 \pm ,027$ & $.704 \pm .021$ & $.890 \pm .008$ \\ \hline
\multicolumn{1}{c|}{\multirow{3}{*}{\begin{tabular}[c]{@{}c@{}}DD-R-DRO\\
(relative)\\ $\alpha = 90\%$\end{tabular}}} & Train & $.030\pm .014$ & $.244
\pm .121$ & $.375 \pm .038$ & $.452 \pm .067$ & $.402\pm .058$ & $.234 \pm
.032$ \\ 
\multicolumn{1}{c|}{} & Test & $.141 \pm .054$ & $.300\pm $.108 & $.556 \pm
.022$ & $.577 \pm .032$ & $.610\pm .024$ & $.332 \pm .011$ \\ 
\multicolumn{1}{c|}{} & Accur & $.949 \pm .019$ & $.921 \pm .070$ & $.729
\pm .023$ & $.717\pm .025$ & $.710 \pm .020$ & $.892 \pm .007$ \\ \hline
\multicolumn{1}{c|}{\multirow{3}{*}{\begin{tabular}[c]{@{}c@{}}DD-R-DRO\\
(relative)\\ $\alpha = 50\%$\end{tabular}}} & Train & $.031\pm .016$ & $.232
\pm .094$ & $.445 \pm .032$ & $.544 \pm .057$ & $.365 \pm .054$ & $.288 \pm
.029$ \\ 
\multicolumn{1}{c|}{} & Test & $.154 \pm .049$ & $.319 \pm .078$ & $.570 \pm
.019$ & $.594 \pm .018$ & $.624 \pm .018$ & $.357 \pm .008$ \\ 
\multicolumn{1}{c|}{} & Accur & $.948 \pm .019$ & $.918\pm .081$ & $.705 \pm
.023$ & $.699 \pm .028$ & $.698 \pm .018$ & $.881 \pm .005$ \\ \hline
\multicolumn{2}{c|}{Num Predictors} & $30$ & $4$ & $30$ & $10$ & $20$ & $56$
\\ 
\multicolumn{2}{c|}{Train Size} & $40$ & $20$ & $80$ & $30$ & $30$ & $150$
\\ 
\multicolumn{2}{c|}{Test Size} & $329$ & $752$ & $475$ & $9990$ & $125034$ & 
$2951$%
\end{tabular}
  }

  \label{Tab-real}
\end{table}

We observe that the doubly robust DRO framework, in general, get robust
improvement comparing to its non-robust counterpart with $\alpha =90\%$.
More importantly, the robust methods tend to enjoy the variance reduction
property due to RO. Also, as the robust level increases, i.e. $\alpha =50\%$%
, where we believe in higher noise in cost function learning, we can
observe, the doubly robust based approach seems to shrink towards to LRL1,
and  benefits less from the data-driven cost structure.

\section{Discussion and Conclusion}

We have proposed a novel methodology, DD-R-DRO, which calibrates a
transportation cost function by using a data-driven approach based on RO. In
turn, DD-R-DRO uses this cost function in the description of a DRO
formulation based on optimal transport uncertainty region. The overall
methodology is doubly robust. On one hand, DD-DRO, which fully uses the
training data to estimate the underlying transportation cost enhances
out-of-sample performance by allowing an adversary to perturb the data
(represented by the empirical distribution) in order to obtain bounds on the
testing risk which are tight. On the other hand, the tightness of bounds might come at the cost of potentially introducing noise in the testing error performance. The
second layer of robustification, as shown in the numerical examples,
mitigates precisely the presence of this noise.  

\bibliographystyle{apalike}
\bibliography{Double_Robust_DRO}
	
\end{document}